# LP-SLAM: Language-Perceptive RGB-D SLAM system based on Large Language Model

*Weiyi Zhang, Yushi Guo, Liting Niu,*
*Peijun Li, Chun Zhang\*,*
*School of Integrated Circuits,*
*Tsinghua University*
*Haidian, Beijing, China*
{ *wy-zhang20, gys22, nlt22, lpj20*}@mails.tsinghua.edu.cn
{*zhangchun*}@tsinghua.edu.cn

*Zeyu Wan, Jiaxiang Yan, Fasih Ud Din Farrukh,*
*Debing Zhang,*
*School of Integrated Circuits,*
*Tsinghua University*
*Haidian, Beijing, China*
{ *wanzy22, yan-jx21, fa-s17*}@ mails.tsinghua.edu.cn
{*zhangdebing*}@tsinghua.edu.cn

*Abstract* –Simultaneous localization and mapping (SLAM) is a critical technology that enables autonomous robots to be aware of their surrounding environment. With the development of deep learning, SLAM systems can achieve a higher level of perception of the environment, including the semantic and text levels. However, current works are limited in their ability to achieve a natural-language level of perception of the world. To address this limitation, we propose LP-SLAM, the first language-perceptive SLAM system that leverages large language models (LLMs). LP-SLAM has two major features: (a) it can detect text in the scene and determine whether it represents a landmark to be stored during the tracking and mapping phase, and (b) it can understand natural language input from humans and provide guidance based on the generated map. We illustrated three usages of the LLM in the system including text cluster, landmark judgement, and natural language navigation. Our proposed system represents an advancement in the field of LLMs based SLAM and opens up new possibilities for autonomous robots to interact with their environment in a more natural and intuitive way.

## I. INTRODUCTION

Simultaneous Localization and Mapping (SLAM) is a perception-based problem in robotics that involves constructing a map of an unknown environment while simultaneously determining the robot's position in real-time. Traditional SLAM algorithms rely on sensors such as radars[1] and cameras [2][3][4] to build a geometric model of the environment and estimate the robot's pose.

The need to handle extraordinarily dynamic or complex environments has led to the development of semantic SLAM[5][6], where semantic information such as object-specific categories is used to improve perceptiveness. Deep learning techniques such as R-CNN[7], Faster R-CNN [8], and Mask RCNN [9] have given rise to neural network-based semantic SLAM, which is utilized to obtain the semantic information in the environment. Work [10] applied a CNN to segment 3D points and map semantic labels to objects using a nearest-neighbor approach, based on which to update the map of target object point cloud information and subordinate type confidence value. In work [11], a semantic SLAM system called DS-SLAM was proposed, which leverages the SegNet architecture [12] for performing real-time pixel-level semantic segmentation using the caffe framework [13]. DS-SLAM not only applies semantic information to mapping and object recognition, but also utilize it to filter outliers in the process of tracking in a dynamic environment. In addition, semantic information was used to estimate the camera pose in work [14]. In Work [15], an object co-view was constructed with the semantic information for checking loop candidates based on the underlying geometric features during the loopback detection phase. In the above works, semantic offers high-level feature information that improves localization accuracy, masks dynamic feature points, and assists in bundle adjustment (BA) and loopback detection in SLAM.

Text is another type of complex feature that can be extracted by SLAM. Several attempts have been made to use text for navigation and location. Work [16] built a text-assisted visual inertial SLAM system for blind people. Work [17] used SLAM to extract planar tiles, integrated multiple observations to detect text, then fused the consecutive detections. In 2015, the same team proposes "junction" descriptor for text spotting, integrated into real-time SLAM, which improves the localization and text identification [18]. Work [19] developed a text-based visual SLAM method that uses detected text as a planar feature with three parameters and illumination-invariant photometric error, resulting in more accurate 3D text maps for robotic and augmented reality applications.

Despite the promise of semantic or text-based SLAM, there are still challenges to be addressed. Current semantic SLAM systems are unable to utilize the information from texts, and the SLAM systems based on texts focus on the special planar geometry information of texts rather than the actual meaning of texts. In addition to the geometry-level SLAM systems based on planar geometry features and semantic-level SLAM systems with semantic perception, we propose the first natural-language-level SLAM named LP-SLAM leveraging the emerging large language models [20][21][22][23]. LP-SLAM leverages OCR technology to recognize image features at the natural language level and integrates the language feature into visual SLAM using LLMs. The contributions of LP-SLAM include:

1) As the first natural-language-level SLAM system, LP-SLAM has the ability of language perception in three major aspects: single text judgement, multiple texts clustering, and natural-language-driven guidance for navigation. The detected texts judged as landmarks will be stored in the map during the mapping phase. Given the users' requirements in natural language, the SLAM system tells the user where to go to accomplish the demand.

2) The representative model of LMMs, Chat-GPT, is introduced into the LP-SLAM to process the text information extracted from a deep-neural-network-based scene text recognition (STR) module during the mapping phase. Chat-GPT is also used to bridge the users' natural language and the data of SLAM system for navigation. We demonstrate how the ChatGPT is used as three different key functions in the system.

3) Techniques inspired by human cognition are introduced to deal with the STR mis-detection and mis-recgonition cases. The similarity classification strategy is designed for higher robustness against the mis-recognition of the text. A human-conception inspired long-short-term memory strategy is designed to deal with the mis-detection cases and reduce the calculation.

4) We conduct experiments in the environment simulating the scene of super mall, where the key landmarks are the shop names. The results show our LP-SLAM has the potential to enhance autonomous robots' ability to interact with their environment in a more natural and intuitive way, offering a promising avenue for future research.

## II. BACKGROUND KNOWLEDGES

### A. Visual SLAM

Simultaneous Localization and Mapping (SLAM) has emerged as a significant research topic in the field of robotics and computer vision. SLAM techniques enable a robot to navigate in an unknown environment by building an incremental map of the surroundings while simultaneously estimating its own movement. This approach is particularly critical in scenarios where prior knowledge of the environment is not available and crucial in the realization of fully autonomous mobile robots.

Visual SLAM, a variant of SLAM, utilizes vision as the sole external sensory perception method. MonoSLAM [24] was the first real-time visual SLAM system that employed a monocular camera and Extended Kalman Filter (EKF) to estimate camera location and construct a sparse 3D map. However, the high computational complexity and limited accuracy of the filtering algorithm hindered further development. To address these issues, PTAM [25] was proposed as the first SLAM system that utilized graph optimization instead of filtering. PTAM's keyframe approach achieved superior performance with lower cost, and it introduced the concept of front-end tracking and back-end optimization to improve efficiency.

ORB-SLAM[26], which used FAST corner points and Oriented Brief (ORB) descriptors, improved running speed and accuracy compared to PTAM. Additionally, ORB-SLAM incorporated Loop Closure Detection to optimize the map's pose and reduce accumulated drift errors. The Bag of Words library DBoW2 was also introduced to improve feature matching and loop closure detection speed. ORB-SLAM2[2] extended the framework to support binocular and RGBD cameras, while also adding a pre-processing module in the tracking thread to handle more information from these devices, thus improving precision. Furthermore, it introduced full BA to rectify the map and added a localization mode, which disabled local mapping and loop closing threads, allowing the camera to relocate through the tracking thread. ORB-SLAM3[27] is the latest release of the series and supports an even wider range of equipment and functions, such as pinhole, fisheye, and visual inertia odometer. It is a multi-map system that creates a new map when VO is lost, and automatically merges with the previous map when the scene is recovered. ORB-SLAM3 has demonstrated robustness comparable to state-of-the-art systems while achieving higher precision. ORB-SLAM series have become representative modern SLAM system, and subsequent research has been carried out based on its open-source project.

SLAM systems using direct methods were also developed. LSD-SLAM[28] is a type of direct SLAM based on optical flow tracking, which realizes semi-dense map construction and can run online in real-time. DSM[29] is the first SLAM system that realizes loopback detection and map reuse entirely based on a direct method and achieves the highest accuracy of direct method SLAM on several datasets.

### B. Scene text detection

Scene text detection and recognition are crucial tasks in computer vision, and Optical Character Recognition (OCR) technology is widely used in different fields. In this part, we present an overview of some of the most commonly used algorithms for scene text detection and recognition.

CTPN [30] is a popular method for text detection, which directly detects text lines in a sequence of fine-scale text proposals within convolutional feature maps. It utilizes a vertical anchor mechanism to jointly predict the location and text/non-text score of each fixed-width proposal, resulting in improved localization accuracy. EAST [31] predicts words or text lines of arbitrary orientations and quadrilateral shapes in full images, eliminating unnecessary intermediate steps such as candidate aggregation and word partitioning with a single neural network. DBNet [32] is a segmentation-based method that better deals with the irregular shape of text such as bending. It proposes a Differentiable Binarization (DB) module that simplifies the post-processing process by adding the binary threshold to the training and obtaining a more accurate detection boundary.

Moving to scene text recognition, CRNN [33] utilizes mainstream convolutional structures for feature extraction, while text recognition algorithm introduces Long-Short-Term Memory (bidirectional LSTM) to enhance context modeling. The output feature sequence is then input to the CTC module, and the sequence result is decoded directly. ASTER [34] comprises a rectification network and a recognition network, and the recognition network is an attentional sequence-to-sequence model that predicts a character sequence directly from the rectified image. SRN [35] is a novel end-to-end trainable model that uses a global semantic reasoning module (GSRM) to capture global semantic context through multi-way parallel transmission. This module enables accurate scene text recognition by enhancing context modeling.

## C. Large Language Model

A large language model (LLM) is a type of artificial intelligence (AI) system that is designed to process natural language and generate coherent, contextually appropriate responses to user inputs. LLMs are typically based on deep learning algorithms that are trained on massive datasets of text, allowing them to recognize patterns and relationships in language and generate highly accurate and relevant responses.

The development of Large Language Models has been a gradual process that has spanned several decades. The earliest models were simple rule-based systems that relied on hand-crafted linguistic rules to process text. However, as machine learning and deep learning techniques have become more advanced, Large Language Models have become increasingly sophisticated. One of the earliest examples of a large language model was the Statistical Language Model (SLM) [36], which was developed in the 1990s. This model was based on statistical techniques such as n-grams [37] and maximum likelihood estimation, and was used for tasks such as speech recognition and machine translation.

The most popular Large Language Models in use today are based on the Transformer architecture [38], which was introduced in 2017 by researchers at Google. The Transformer architecture is a neural network architecture that was proposed by Vaswani et al. in 2017. It was designed to overcome some of the limitations of traditional recurrent neural networks (RNNs) and convolutional neural networks (CNNs) in processing sequential data, such as human language. The Transformer architecture incorporates the self-attention mechanism, which allows the model to attend to various parts of the input sequence and capture complex relationships between them. The Transformer architecture offers several advantages over traditional RNNs and CNNs. It can process sequences in parallel, which makes it more efficient than RNNs and CNNs. It also enables the model to capture long-range dependencies and relationships between words and phrases, which is essential for natural language processing applications, such as language translation and language modeling.

GPT-3 (Generative Pre-trained Transformer 3) [39] is a Large Language Model developed by OpenAI in 2020, based on the Transformer architecture. It is one of the largest and most powerful language models to date, with 175 billion parameters. GPT-3 was trained on a massive corpus of text data, including web pages, books, and articles, and can generate highly coherent and fluent text.

One of the key differences between GPT-3 and other transformer-based models is its ability to perform few-shot and zero-shot learning. Few-shot learning refers to the ability to learn from a small amount of labeled data (e.g., a few examples of a task), while zero-shot learning refers to the ability to perform a task without any labeled data. GPT-3 achieves this by leveraging its large language model and using prompt-based learning, where the user provides a prompt (e.g., a few example sentences) and the model generates the desired output.

Another important feature of GPT-3 is its ability to perform multitask learning, where the model is trained on multiple NLP tasks simultaneously. This allows GPT-3 to generalize better across different tasks and domains.

GPT-3 has demonstrated remarkable performance on a wide range of natural language processing tasks, such as language understanding, language generation, and language translation. It can generate text in various styles and tones, and can even answer questions and complete sentences. Its advanced capabilities have significant implications for the field of natural language processing and offer exciting opportunities for applications in areas such as chatbots, content generation, and language translation.

## III. PROPOSED METHOD

### A. Overall framework

The overall framework of LP-SLAM is shown in Fig.1. The LP-SLAM introduces three language-processing workflows based on ORB-SLAM2: runtime text mapping, distilling, and navigation. Three ChatGPT-assisted modules are marked in red.

The runtime text mapping thread is executed along with the tracking thread of the SLAM system. The input RGB image is firstly processed by the deep neural network (DNN) based scene text recognition (STR) module to extract the set of visible texts $T_i = t_{i1}, t_{i2}, ..., t_{in}$, where $t_{in}$ denotes the n-th text in the i-th image. One text may appear in many frames and the extraction results differ in frames due to the limited accuracy of STR. Thus, the similarity module is implemented to classify the words that may extracted from the same text in the real world, which is called homologous texts. For example, the brand name "Alienware" may be extracted as "Allenware", "Alienvvare", and "Aiienware". The similarity classification module recognizes the homologous texts and merges them into one class for the following processing. In addition to the mis-recognizing of one text, there may be cases where a non-text object or texture is mistakenly detected as one text. This often occurs as that one logo is recognized as one character such as "A", or that one decorative pattern is recognized as a series of meaningless characters. We introduce the long-short-memory module to avoid the mis-recognizing inspired by human's short-term visual memory and long-term memory of brain. At the consumption in that the mis-recognized texts only appear for very few times, only the correct detected texts are stored into long-term memory and processed further. Each class of homologous texts will then go through the long-short-term memory module. Once one class of homologous texts enters the long-term memory, the ChatGPT based text cluster is performed to select the most representative and reasonable name of the class. For example, the correct brand name "Alienware" is selected from the mis-spelling texts as the final name of the class. The position of the current viewed text in the long-term memory is calculated based on the camera pose $T_{wc}$ and the pixel position of the text in the RGB image. One text appears in several images in general, thus the positions calculated from different frames are all stored for the later position cluster.

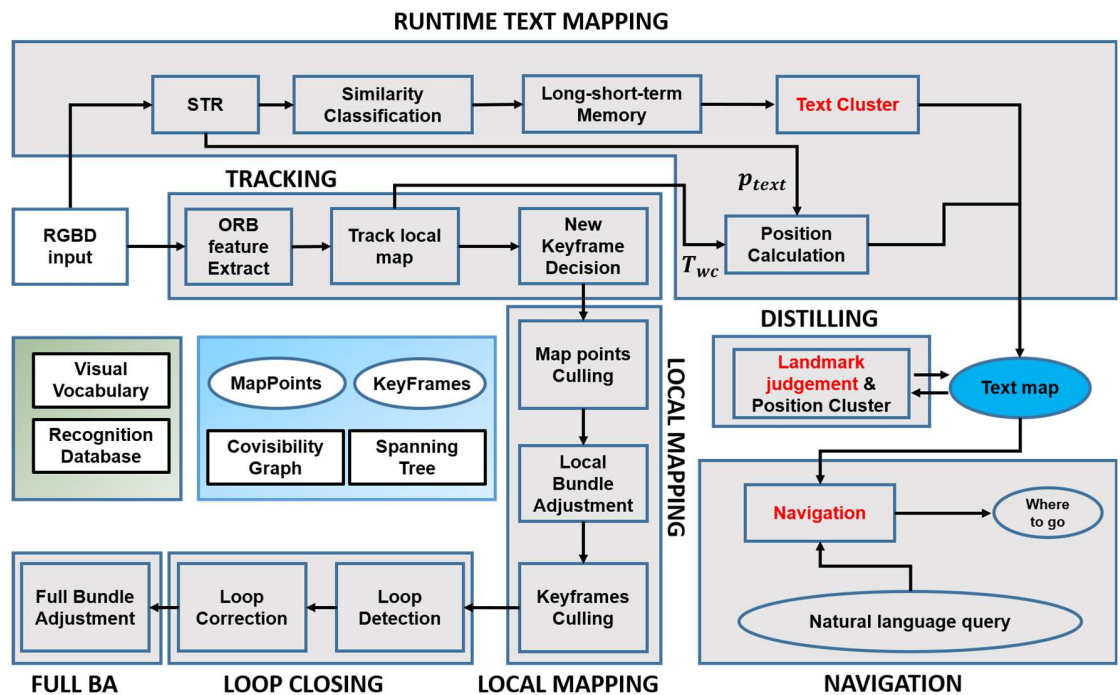

Fig.1 Overall framework of LP-SLAM.

The distilling workflow, containing two functions, is executed along with or after runtime text mapping. The classes of homologous texts are judged by ChatGPT to decide whether they are names of landmarks. The texts that fail to pass the judgement are removed from the long-term memory. Position cluster is performed on the reserved items to generate the corresponding positions in the world coordinates by clustering algorithms from the stored positions.

The navigation workflow is executed after the above two workflows when the text map and point cloud map are established. The query from users in natural language is input and processed by ChatGPT to understand the users' demand. Then ChatGPT selects the suitable landmark as the target and SLAM system returns the corresponding position, which can be used for the follow-up navigation algorithms.

*B. Scene text recognition*

The scene text detection and recognition are implemented by OpenCV. The DBNet algorithm is utilized for scene text detection, while the CRNN algorithm is utilized for text recognition. Different from traditional methods relying on a threshold to convert the segmentation probability graph into a binary graph, DBNet uses a binary approximate function, DB, which is fully differentiable in network training. The DB model is effective in dealing with irregular text shapes such as bending and is capable in both Chinese and English detection. CRNN is primarily used for end-to-end recognition of indefinite length text sequences. The model comprises convolutional layers, which extract features from the input image to obtain the feature map, recurrent layers that use bidirectional RNN (BLSTM) to predict the feature sequence and output the prediction label distribution, and transcription layers, which use a CTC loss to convert a series of tag distributions taken from the loop layer into a final tag sequence. During the processing of input image, the detection model is first executed to locate the scene text, and the positions of extracted texts are stored. Then the text boxes are rectified into rectangular boxes by four-point perspective transformation. Finally, the rectified image boxes patches are processed by the recognizing module to generate the texts.

*C. Similarity classification*

The STR may identify minor errors in the characters of some shop names, resulting in the potential for multiple names to refer to the same shop. To address this issue, it is crucial to determine the level of similarity between the strings and group those that belong to the same shop into clusters. In this paper, the Levenshtein Distance algorithm is proposed to implement this function, which ensures that names with sufficient similarity are retained for categorisation and those with insufficient similarity are eliminated. The Levenshtein distance algorithm is a dynamic programming algorithm that measures the similarity between two strings by calculating the minimum number of insertions, deletions and substitutions required to convert them. This algorithm employs the concept of backtracking in the comparison process, allowing for recursive operations. To carry out the comparison between two strings A and B of lengths m and n, respectively, a matrix $D[n + 1][m + 1]$ is constructed. This matrix is populated by circulating through each cell $D(i, j)$ and calculating its corresponding value. The state transfer equation is as follows:

$$D(i,j) = \begin{cases} 0, & i = 0, j = 0 \\ j, & i = 0, j > 0 \\ i, & i > 0, j = 0 \\ Min, & i > 0, j > 0 \end{cases} \quad (1)$$

$$Min = 1 + min \begin{cases} D(i-1)(j), \\ D(i)(j-1), \\ D(i-1)(j-1) - k \end{cases} \quad (2)$$

where $k = 1$ when the last letters of the two strings are the same else $k = 0$. After recursive calculation, the final $D(n+1, m+1)$ is the final edit distance length. We utilize equation (3) to normalize the edit distance to obtain a more accurate measure of string similarity, regardless of differences in length. This allows us to establish a uniform threshold for classification. If two strings are deemed sufficiently similar, they will be grouped into a single category, ready for processing by chatGPT.

$$Sim(A, B) = 1 - \frac{D(n+1, m+1)}{\max(m, n)} \quad (3)$$

*D. Long-short-term memory*

The volume of data collected from the OCR is substantial, comprising three types of information: accurate data at high frequencies, slightly erroneous data at medium frequencies, and irrelevant data at low frequencies. Drawing inspiration from the process by which the human brain differentiates memories based on the intensity of visual signals, we introduce the long-short-term memory strategy to effectively eliminate mis-identified text and integrate medium and high frequency data.

Information that enters the short-term memory stage is first processed by the similarity classification module proposed above. This module combines correct information and slightly erroneous data into classes, upon which subsequent memory processes depend. When a text is assigned to a specific class, the memory score for that class increases. To prevent score overflow, we have implemented a function that simulates the human brain's forgetting mechanism. This function reduces the memory score of all classes in each frame. Once the memory score for a given category reaches the predetermined magnitude, it is transferred to the long-term memory, which can later be used for categorization and navigation tasks. The overall process of the long-short-term memory strategy is shown in Fig.2. It shows the information passing through the similarity classification module is in terms of classes to stimulate the intensity of short-term memory and the long-term and short-term memories are stored in separate spaces. In this example, it's evident that class1 with a magnitude above the threshold is stored in the long-term memory pool. This mechanism ensures that low-frequency error information is never transferred to long-term memory, thereby eliminating the influence of irrelevant data on the system from the outset.

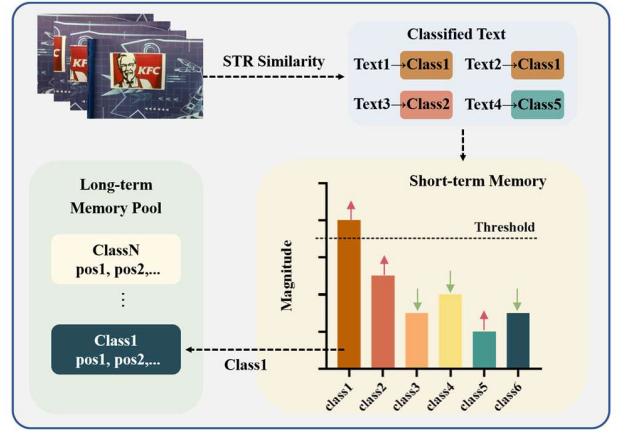

Fig.2 Workflow of long-short-term memory

*E. Text clustering*

OCR errors occur due to a variety of factors, such as image noise, distortion, and complexity of the neural network, as well as insufficient training data. For instance, when presented with a video of the word "HIGHTEMPERATURE," OCR may produce variations such as "HIG/eMPERAURe" or "PNIGHIYEMIPERATUN," containing spelling errors. Although a long-short-term memory system could potentially group such variations together, the issue of selecting the correct word remains, which is commonly referred to as the "text cluster problem."

Traditional algorithms may involve constructing a large dictionary that encompasses every possible word in every language, but this approach is time-consuming due to the need for extensive searching. Another method is to train a massive neural network to identify meaningful words; however, this approach is also limited, as the error rate increases when the number of words within a given landmark increases.

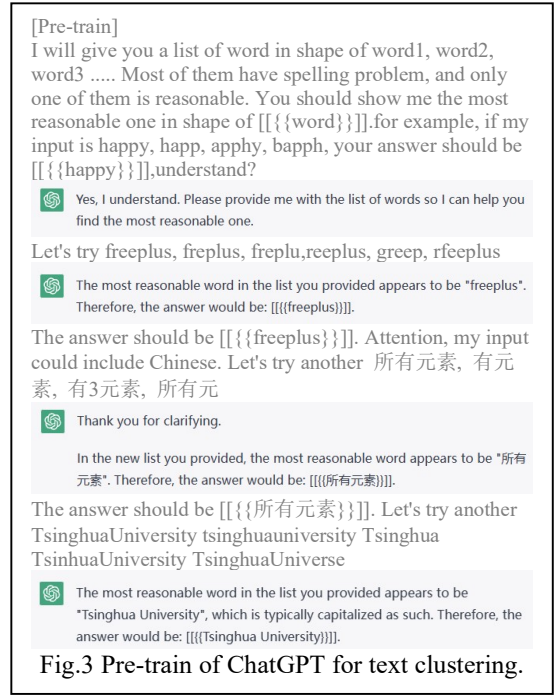

Fig.3 Pre-train of ChatGPT for text clustering.

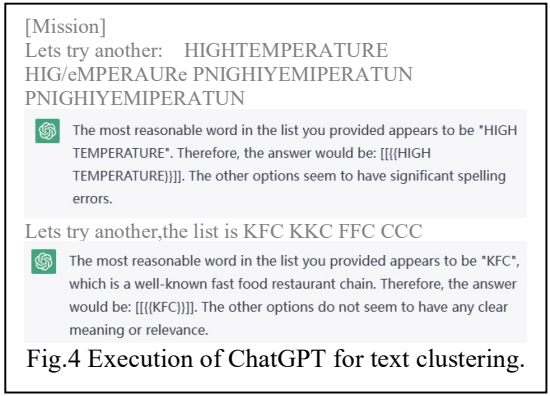

Fig.4 Execution of ChatGPT for text clustering.

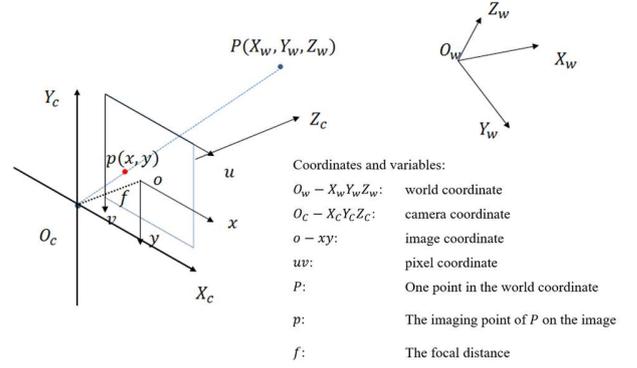

Fig.5 Transform between coordinates.

To address these challenges, ChatGPT presents a promising solution. By providing ChatGPT with a set of similar words, it can identify the most appropriate word within the group. To effectively leverage ChatGPT for text clustering, a pipeline is constructed that involves the following steps:
1. First, the long-short-term memory system generates a set of similar words, many of which contain spelling errors.
2. ChatGPT is pre-trained on this word list to identify the most meaningful word within the given set.
3. Once pre-training is completed, the pipeline can be extended to include words with spelling problems in multiple languages. In this scenario, ChatGPT is capable of selecting the most appropriate word from the list provided, thereby addressing the text cluster problem.

The pre-training process also consists two parts, mission description and example training. Fig.3 illustrates the pre-training process, while Fig.4 outlines how users interact with ChatGPT.

### F. Position calculation

The position of the viewed text in the world coordinates is calculated based on the pixel position from STR and the current camera pose from SLAM tracking thread. Fig.5 illustrates the transform relationship between the real point $P(x_w, y_w, z_w, 1)$ in the world coordinate and the corresponding image point $(u, v)$. The real point $P(x_w, y_w, z_w)$ is firstly transformed into camera coordinate as:

$$P(x_C, y_C, z_C, 1) = T_{CW} * P(x_w, y_w, z_w, 1) \quad (4)$$

where $T_{CW}$ is the 4x4 transformation matrix from world coordinate to camera coordinate. Then $P(x_C, y_C, z_C, 1)$ is projected onto the image coordinates as:

$$\begin{bmatrix} x \\ y \\ 1 \end{bmatrix} = \frac{1}{Z_c} * \begin{bmatrix} f_x & 0 & 0 & 0 \\ 0 & f_y & 0 & 0 \\ 0 & 0 & 1 & 0 \end{bmatrix} * \begin{bmatrix} x_c \\ y_c \\ z_c \\ 1 \end{bmatrix} \stackrel{\text{def}}{=} \frac{1}{Z_c} * F_c * \begin{bmatrix} x_c \\ y_c \\ z_c \\ 1 \end{bmatrix} \quad (5)$$

Finally, the pixel position is calculated by affine transformation based on the intrinsic of the camera:

$$\begin{bmatrix} u \\ v \\ 1 \end{bmatrix} = \begin{bmatrix} \frac{1}{d_x} & 0 & u_0 \\ 0 & \frac{1}{d_y} & v_0 \\ 0 & 0 & 1 \end{bmatrix} * \begin{bmatrix} x \\ y \\ 1 \end{bmatrix} \stackrel{\text{def}}{=} I_c * \begin{bmatrix} x \\ y \\ 1 \end{bmatrix} \quad (6)$$

Inversely, the position $P$ in real world coordinate can be calculated from the pixel position $(u, v)$ as:

$$P(x_w, y_w, z_w, 1) = Z_c * T_{wc} * F_c^{-1} I_c^{-1} * \begin{bmatrix} u \\ v \\ 1 \end{bmatrix} \quad (7)$$

The pixel position of the detected text is calculated as the average of the four counters of the detection box.

### G. Landmark judgement

There are various street signs that act as landmarks, such as "STOP" or "Supermarket." When we are hungry, our eyes gather information from these landmarks, and our brain identifies those that represent restaurants and locates their position. This process is known as landmark judgment, which is used to classify landmarks based on their language-perceptive characteristics.

Traditional semantic slam methods, such as DS-SLAM, utilize a semantic thread to extract information about the environment but do not take into account the textual information on the landmarks. Neural networks can transform landmarks into semantic labels, but these networks are trained by engineers, and users are unable to request labels that engineers have not yet offered. Therefore, we propose landmark judgment realized by ChatGPT. Trained on a large amount of data, ChatGPT is capable of fulfilling almost all natural language requests related to the environment as long as we provide information about the landmarks. To effectively utilize ChatGPT for landmark judgment, we construct a pipeline consisting of the following steps:
1. First, after conducting a scan of the surroundings with SLAM, we extract all the landmarks and generate them into a list of words and their address using OCR.
2. Next, we pre-train the ChatGPT agent to ensure that its responses are precise and straightforward to extract.
3. Once pre-trained, the users' queries will be forwarded to ChatGPT, which will provide an accurate and easy-to-process answer.

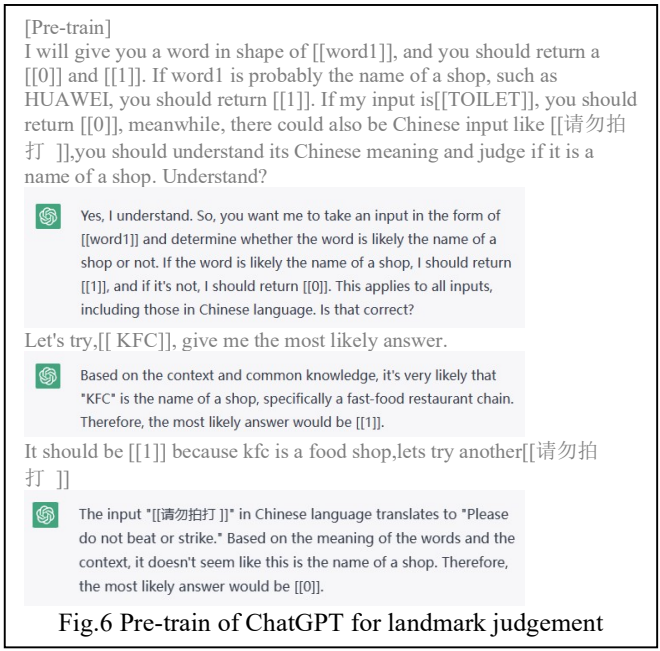
Fig.6 Pre-train of ChatGPT for landmark judgement

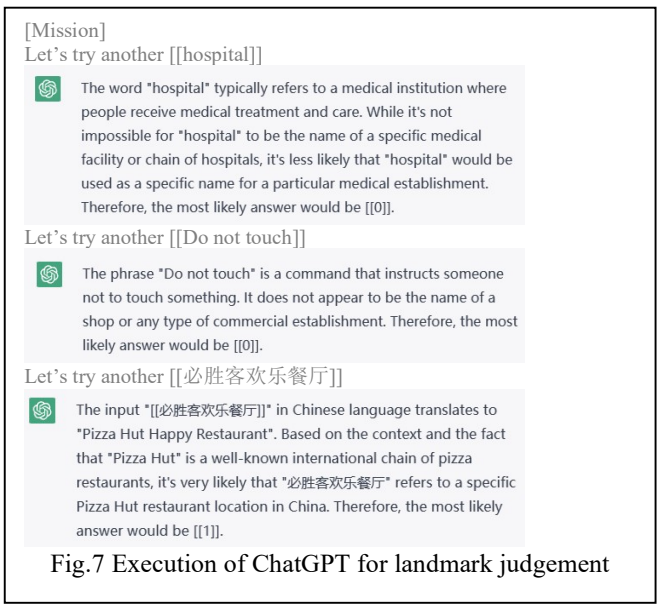
Fig.7 Execution of ChatGPT for landmark judgement

The pre-training process consists of two key components. Initially, we introduce ChatGPT to the task it is expected to accomplish and subsequently provide it with relevant examples to ensure optimal performance. Fig.6 illustrates the pre-training process, while Fig.7 outlines how users interact with ChatGPT.

### H. Position clustering

The clustering method is implemented to generate the final position of each class of homologous texts. For each class, $N$ iterations of clustering are executed to remove the outliers. In each iteration, the average position $\bar{p}$ is calculated, which is also the optimum of least squares. Then the 20% farthest points from the average position are judged as outliers and removed.

### I. Navigation

Conventional navigation systems rely on precise location data such as latitude and longitude or street addresses. While contemporary navigation software has the capability to understand user requests like "where is the nearest restroom," users must simplify their language to a specific format that the software can comprehend. Additionally, the process of training neural networks for such tasks is resource-intensive and time-consuming.

The advent of ChatGPT provides a solution to these challenges by allowing for natural language processing during navigation without the need for additional network training. However, due to the natural language-like output generated by ChatGPT, pre-training is a requisite step to make the output manageable. The following is the pipeline for ChatGPT-based navigation:

1. First, pre-train ChatGPT to only provide coordinates for navigating to specific location.
2. We present ChatGPT with a list of structures containing places and their respective coordinates, which can be generated using SLAM.
3. Users can then issue natural language commands to ChatGPT, which will return a specific location to navigate towards.

Fig.8 shows how we pre-train ChatGPT, and Fig.9 shows how the users interact with ChatGPT.

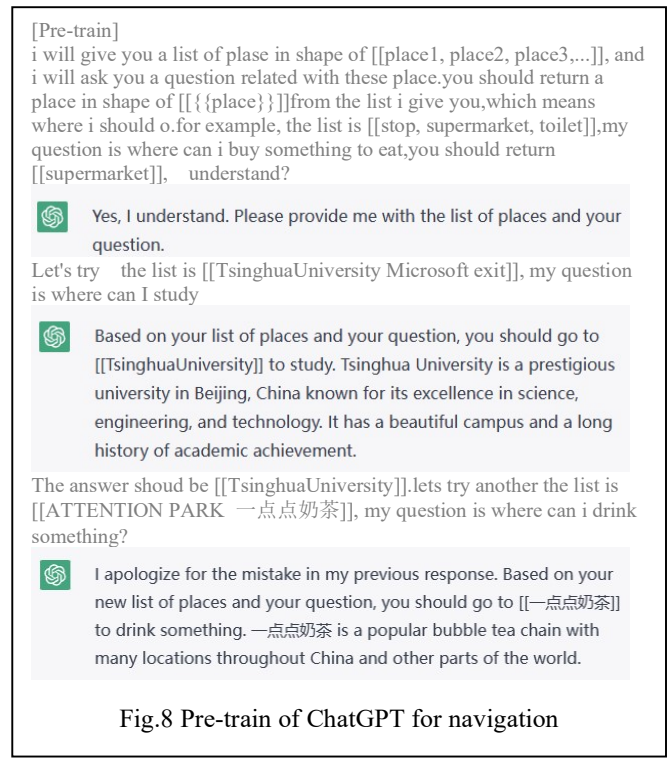
Fig.8 Pre-train of ChatGPT for navigation

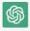

Fig.9 Execution of ChatGPT for navigation

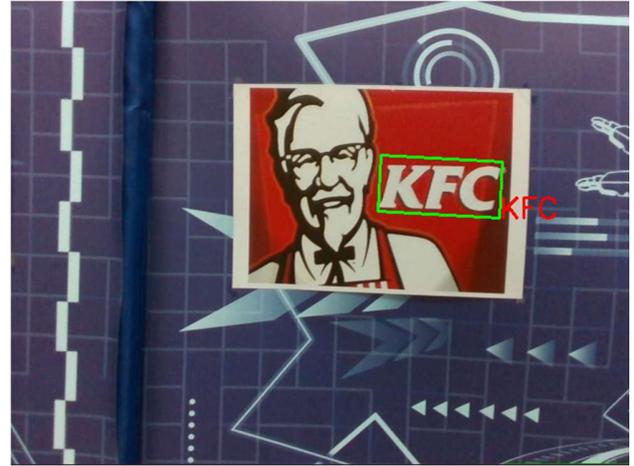

Fig11. (a) Correctly detected case

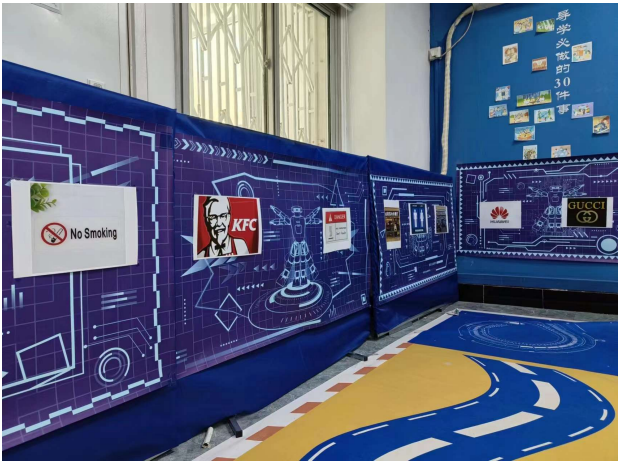

Fig.10 Experimental environment of the mock mall.

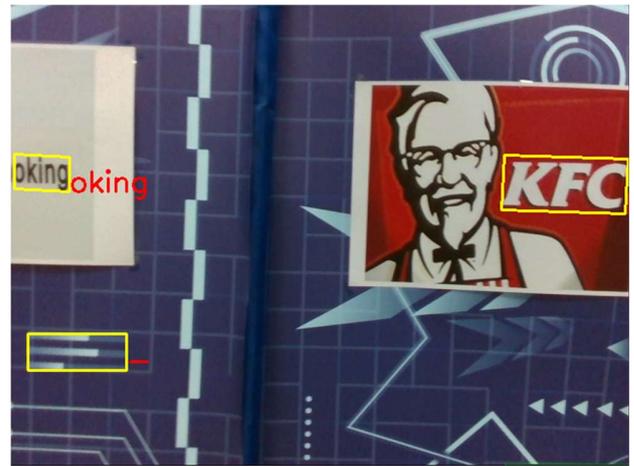

Fig11. (b) Mis-detecting and out-of-thread cases

## IV. EXPERIMENTAL RESULTS

The experiment is conducted in a mock mall environment containing shops, slogans, and public facilities, as shown in Fig.10. The platform robot is Roban child-sized humanoid robot from Leju Robotics with Intel RealSense D435i RGBD camera.

*A. Scene text recognition*

The result of scene text recognition is illustrated as Fig. In Fig.11 (a), the brand name "KFC" is correctly detected and recognized. The pixel position of the text is also within the threshold; thus, the detected text is surrounded by green box and is reserved for the following processes. In Fig.11 (b), three texts are detected and recognized. However, they are all surrounded by yellow box and will be eliminated. The left upper one is actually one part of the slogan "No Smoking", which is not fully viewed at the current image. Cases where texts are not fully viewed will result in out-of-threshold edges of detected box. In this example, the left boundary of the detected box is too far to the left and reaches out the threshold. Thus, it is eliminated. The left bottom text is one case of mis-detecting, which means that one non-text area is judged as text-area by the detecting module. In this image, it will be eliminated because of the out-of-range location. However, in other cases where mis-detecting happens in the threshold, the following processes will further filter out the outliers. The correctly detected text "KFC" is also eliminated with the same reason. Based on the experiments, a larger threshold rejects more outliers and thus improves the robustness of the system. The implemented text recognition gets mistaken results at cases. For example, "请勿拍打" (Do not beat in English) is always recognized as "请切拍打" (Meaningless Chinese sentence). The following techniques will prevent the performance decrease from such cases.

*B. Similarity classification*

Fig.12 illustrates the outcomes of the similarity classification module processing. On the left side, the data received before processing is shown, which is disorganized and contains errors. If this data was kept separately, it would have a negative impact on both resources and navigation accuracy. Therefore, the similarity classification module is utilized to determine string similarity, identifying which data to keep and which to discard as meaningless. The number preceding each character indicates its frequency, which is an essential basis for subsequent long-short-term memory mechanisms. In Figure Xb, the medium and high-frequency meaningful data are highlighted in green, which are grouped together if they meet the similarity threshold. Conversely, the low-frequency meaningless data is in red, which will be discarded in the long-short-term module.

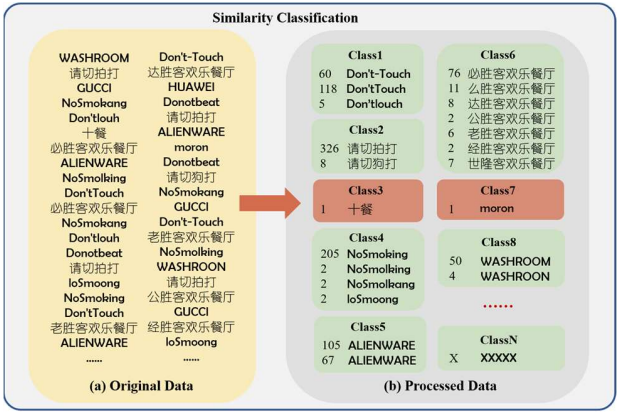

Fig.12 Results of similarity classification.

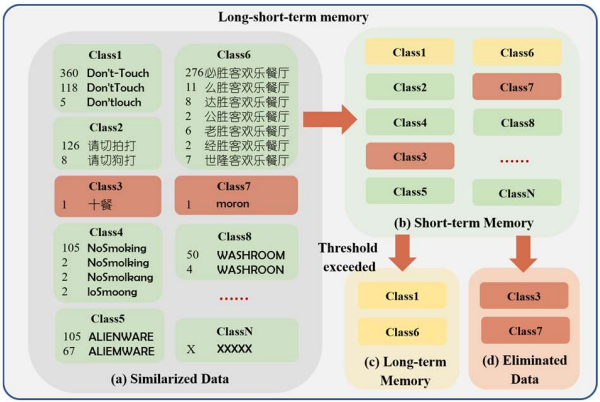

Fig.13 Results of long-short-term memory

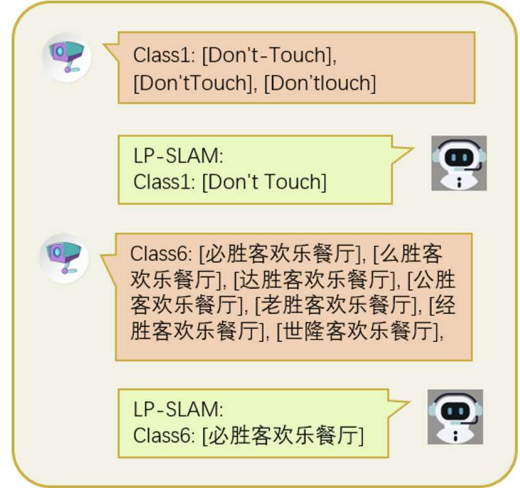

Fig.14 Results of text clustering

*C. Long-short-term memory*

Fig.13 illustrates the outcomes of the long-short-term and short-term memory strategy. Specifically, Fig.13(a) presents the clustering information that was obtained from the SC module, which will undergo further classifying. The memory scores are presented in terms of class statistics, meaning that any string classified as a certain class will impact the memory of that particular class. Fig.13 (b) shows that initially, all classes are stored as short-term memories. However, Class1 and Class6 are classified as long-term memories because they meet the magnitude of threshold. Conversely, Class3 and Class7 are too infrequent and will be eliminated by the module's forgetting mechanism. Through several iterations, the medium and high frequency classes will be stored in the long-term memory pool depicted in Fig.13 (c), while the low frequency and meaningless characters will be eliminated.

*D. Text clustering*

Fig.14 showcases the effectiveness of the Text Cluster function with ChatGPT, which returns the most reasonable landmark name from a group of similar names. In the first dialogue, we present a group of landmark names with spelling variations, including [Don't-Touch], [Dont'tTouch], [Don'tlouch]. ChatGPT is able to accurately identify the most reasonable landmark name, which is [Don't Touch].

The next dialogue demonstrates ChatGPT's ability to handle Chinese landmark names. All the inputs are in Chinese and contain some spelling variations of "Pizza Hut". ChatGPT is able to identify the correct landmark name, despite the variations in spelling. This showcases the versatility of ChatGPT in handling different languages and its ability to accurately cluster similar names.

*E. Landmark judgement*

The Landmark Judgment module in LP-SLAM is responsible for identifying the type of landmark, such as a shop, park, or office building. In our example, we focus on identifying whether a landmark is a shop. To simulate real-world street environments, our dataset includes shop names, warning slogans, and public facilities written in both English and Chinese. As shown in Table.I, LP-SLAM accurately identifies all the landmarks as shops, except for those that are warning slogans, such as "请切拍打( Do not beat)" and "HIGHTEMPERATURE". This showcases the effectiveness of the Landmark Judgment module in accurately identifying different types of landmarks, despite the presence of multiple languages in the dataset.

Table.I Landmark judgement results

| Landmark | Is this a shop? |
|---|---|
| Donotbeat | 0 |
| 请切拍打( Do not beat) | 0 |
| NoSmoking | 0 |
| KFC | 1 |
| Don't Touch | 0 |
| DANGER | 0 |
| HIGHTEMPERATURE | 0 |
| 必胜客欢乐餐厅(pizza hut) | 1 |
| WASHROOM | 0 |
| ALIENWARE | 1 |
| HUAWEI | 1 |
| GUCCI | 1 |

*F. Navigation*

The proposed approach supports multiple languages in landmark names and queries. Fig.15 illustrates how the navigation function works in LP-SLAM. In the first dialogue, after presenting a list of landmarks including "请切拍打" (Do not beat) "Nosmoking" "KFC" "Don't Touch" "DANGER" "HIGHTEMPERATURE" "必胜客欢乐餐厅" (Pizza Hut) "washroom" "ALIENWARE" "HUAWEI" and "GUCCI" the user asks in Chinese, "我不喜欢吃披萨，我应该去哪吃饭?" (I dislike pizza, where can I eat?). LP-SLAM understands the user's request, recommends KFC, and provides its location for easy navigation.

In the second dialogue, when the user asks in English "Where can I eat pizza?", LP-SLAM suggests going to "必胜客欢乐餐厅" (Pizza Hut) in China. This showcases LP-SLAM's ability to handle queries in different languages.

The third dialogue demonstrates LP-SLAM's remarkable ability to handle Japanese queries. When the user asks in Japanese, "ゲームをするためのパソコンはどこで買えますか？" (Where can I buy a computer for games?), LP-SLAM suggests "ALIENWARE" and provides its location for easy navigation.

Overall, LP-SLAM's ability to handle queries in multiple languages provides an added advantage for users who speak different languages or are visiting foreign countries. The integration of ChatGPT into the system further enhances its capability to handle complex queries and provide accurate recommendations.

*G. Overall Mapping*

The overall mapping is illustrated in Fig.16. The recognized texts are shown in the spares point cloud map. The blue texts are the ones judged as landmarks, while the pink ones are judged as others. The visualized results validate the accuracy of text extraction, conception, and localization.

Fig.15 Results of navigation

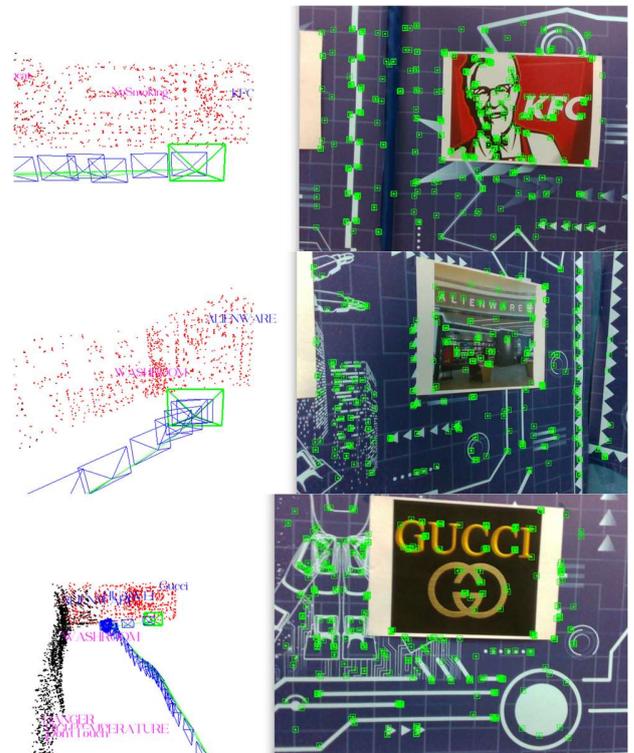

Fig.16 Overall mapping result

V. CONCLUSION

In this work, we propose LP-SLAM, one natural-language-level perception SLAM system leveraging the language understanding capability of LLMs, specifically ChatGPT. LP-SLAM is able to extract the natural-language-level information from the world, judge the information, and store the important information. The human cognition inspired techniques including similarity classification and long-short-term memory are designed to achieve better robustness against mis-detecting and mis-recognition. LP-SLAM is also capable to provide simple navigation guidance according to the query of users in different languages, showing great potential in the real applications. While LP-SLAM illustrates how LLMs improve the perception of SLAM into brand new natural language level, there are still further works to be researched. For instance, topics including how language information can help in accuracy and efficiency of the SLAM threads are potential fields.

ACKNOWLEDGMENT

Supported by the National Natural Science Foundation of China (No.U20A20220).